\documentclass[sn-basic]{sn-jnl}% Basic Springer Nature Reference Style/Chemistry Reference Style
\jyear{2021}%

\theoremstyle{thmstyleone}%
\theoremstyle{thmstyletwo}%

\theoremstyle{thmstylethree}%

\usepackage{booktabs}
\usepackage{soul}
\usepackage{amsmath}
\usepackage{amssymb}
\usepackage{makecell}
\usepackage{bbding}
\usepackage{pifont}
\usepackage{bigstrut}
\usepackage{color}
\usepackage{algorithm}
\usepackage{algpseudocode}
\usepackage{multirow}

\begin{document}

\title[Rethinking Person Re-Identification via Semantic-Based Pretraining]{Rethinking Person Re-Identification via Semantic-Based Pretraining}


\author*[]{\fnm{Suncheng} \sur{Xiang}}\email{xiangsuncheng17@sjtu.edu.cn}

\author[]{\fnm{Jingsheng} \sur{Gao}}\email{gaojingsheng@sjtu.edu.cn}

\author[]{\fnm{Zirui} \sur{Zhang}}\email{654418374@sjtu.edu.cn}

\author[]{\fnm{Mengyuan} \sur{Guan}}\email{gemini.my@sjtu.edu.cn}

\author[]{\fnm{Binjie} \sur{Yan}}\email{yanbinjie@sjtu.edu.cn}

\author[]{\fnm{Ting} \sur{Liu}}\email{louisa\_liu@sjtu.edu.cn}

\author[]{\fnm{Dahong} \sur{Qian}}\email{dahong.qian@sjtu.edu.cn}

\author[]{\fnm{Yuzhuo} \sur{Fu}}\email{yzfu@sjtu.edu.cn}

%\affil[1]{\orgdiv{School of Electronic Information and Electrical Engineering}}
%\affil[2]{\orgdiv{School of Biomedical Engineering}}
\affil[]{\orgname{Shanghai Jiao Tong University}, \orgaddress{\city{Shanghai}, \postcode{200240}, \country{China}}}

%\affil[2]{\orgdiv{School of Biomedical Engineering}, \orgname{Shanghai Jiao Tong University}, \orgaddress{\city{Shanghai}, \postcode{200240}, \country{China}}}

%\affil[1]{\orgdiv{School of Electronic Information and Electrical Engineering}, \orgname{Shanghai Jiao Tong University}, \orgaddress{\city{Shanghai}, \postcode{200240}, \country{China}}}

%\affil[2]{\orgdiv{Department}, \orgname{Organization}, \orgaddress{\street{Street}, \city{City}, \postcode{10587}, \state{State}, \country{Country}}}
%
%\affil[3]{\orgdiv{Department}, \orgname{Organization}, \orgaddress{\street{Street}, \city{City}, \postcode{610101}, \state{State}, \country{Country}}}


\abstract{Pretraining is a dominant paradigm in computer vision. Generally, supervised ImageNet pretraining is commonly used to initialize the backbones of person re-identification (Re-ID) models. However, recent works show a surprising result that CNN-based pretraining on ImageNet has limited impacts on Re-ID system due to the large domain gap between ImageNet and person Re-ID data. To seek an alternative to
traditional pretraining, here we investigate semantic-based pretraining as another method to utilize additional textual data against ImageNet pretraining. Specifically, we manually construct a diversified \textit{FineGPR-C} caption dataset for the first time on person Re-ID events. Based on it, a pure semantic-based pretraining approach named VTBR is proposed to adopt dense captions to learn visual representations with fewer images. We train convolutional neural networks from scratch on the captions of \textit{FineGPR-C} dataset, and then transfer them to downstream Re-ID tasks. Comprehensive experiments conducted on benchmark datasets show that our VTBR can achieve competitive performance compared with ImageNet pretraining – despite using up to $1.4\times$ fewer images, revealing its potential in Re-ID pretraining.}

\keywords{Pretraining, Re-identification, Semantic-based, Visual representation, Convolutional neural network}



\maketitle

\section{Introduction}
\label{sec1}
CNN-based pretraining on ImageNet is a dominant paradigm in computer vision. As many vision tasks are related, it is expected a deep learning model, pretrained on one dataset, to help another downstream task. It is now common practice to pretrain the backbones of object detection~\citep{he2019rethinking} and segmentation~\citep{long2015fully} on ImageNet~\citep{deng2009imagenet} dataset. In the field of person Re-ID, most of works~\citep{xiang2022learning,fu2019self,bai2021unsupervised,zeng2020hierarchical} try to leverage models pretrained on ImageNet to mitigate the shortage of person Re-ID data, which has achieved remarkable performance.
However, this practice has been recently challenged by~\cite{fu2021unsupervised}, who show a surprising result that such ImageNet pretraining may not be the best choice for the re-identification task due to the intrinsic domain gap between ImageNet and person Re-ID data.
Additionally, some research works~\citep{desai2021virtex,radford2021learning} also indicate that learning visual representations from textual annotations can be more competitive to methods based on ImageNet pretraining, which has attracted considerable attention from both the academia and industry worldwide. For these reasons, there has been increasing interests for us to explore novel vision-and-language pretraining strategy which can replace the traditional ImageNet-based pretraining paradigm on Re-ID tasks.
Unfortunately, existing datasets~\citep{zheng2015scalable,ristani2016performance,zheng2017unlabeled,li2014deepreid,xiang2020unsupervised,xiang2021taking} in Re-ID community are all of limited scale due to the costly efforts required for data collection and annotation, especially none of them has diversified attributes to obtain dense captions for pedestrians, which fails to satisfy the need of semantic-based pretraining in Re-ID task.

\begin{figure}[!t]
\centering
\includegraphics[width=1.00\columnwidth]{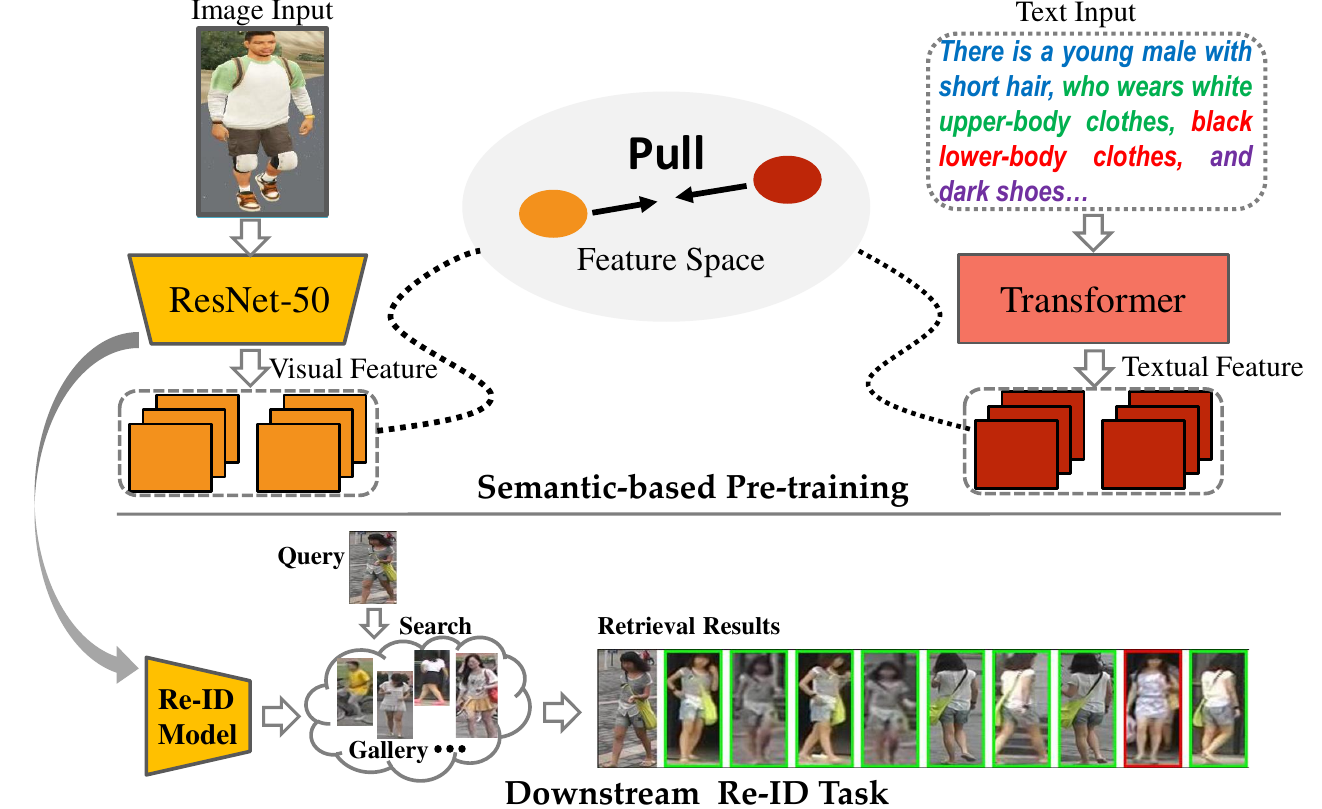}
\caption{The overview of our Re-ID pretraining framework.
First, we jointly train ResNet and Transformer using image caption pairs for the task of image captioning. Then, we transfer the learned ResNet as the backbone of the downstream Re-ID task.}
\label{fig1}
\end{figure}

Targeting to address above mentioned limitations, we start from two aspects, namely data and methodology. From the data perspective, we construct a \textit{\textbf{FineGPR-C}} caption dataset for the first time on person Re-ID events, which involves human describing event in a fine-grained manner. From the methodology perspective, we propose a pure \textbf{V}ir\textbf{T}ex \textbf{B}ased \textbf{R}e-ID pretraining approach named {\textbf{VTBR}, which uses transformers to learn visual representations from textual annotations, the overview of our framework is illustrated in Figure~\ref{fig1}. Particularly, we jointly train a CNN-based network and transformer-based network from scratch using image caption pairs for the task of image captioning. Then, we transfer the learned residual network to downstream Re-ID tasks. In general, our method seeks a common vision-language feature space with discriminative learning constraints for better practical deployment.

%we propose a pure \textbf{V}ir\textbf{T}ex-based \textbf{R}e-\textbf{I}D pretraining approach named VTRI, which uses textual features to learn visual features with fewer images than other approaches. Motivated by this,

%systematically study the problem of Re-ID pretraining, and find that directly applying the commonly used supervised pretrained model on ImageNet or GPR+ does not work well for the Re-ID task.  After a careful investigation

%First, we jointly train a ConvNet and Transformers using image caption pairs for the task of image captioning (top). Then, we transfer the learned ConvNet to several downstream re-ID tasks.

The initial motivation of this research comes from comprehensive study of Re-ID pretraining. In the course of our efforts, we notice that semantic captions can provide a denser learning signal than traditional unsupervised or supervised learning~\citep{desai2021virtex}, so using language supervision on Re-ID task is more appealing, which can provide supervision for learning transferable visual representations with better data-efficiency than other previous approaches.
Another benefit of textual annotation is simplified data collection. Traditional labelling procedure of real pedestrian data always costs intensive human labor, sometimes even involving person privacy concerns and data security problems, which brings researcher a serious challenge for dataset collection. In contrast, natural language description from fine-grained attributes on synthetic data do not require an explicit category and can be easily labelled by non-expert workers, leading to a simplified data labelling procedure without ethical issues regarding privacy.
To the best of our knowledge, we are among the first attempts to use textual features to perform pretraining for downstream Re-ID tasks. We hope this study and  \textit{FineGPR-C} caption dataset will serve as a solid baseline in semantic-based  Re-ID pretraining  and pave a path for community to move forward.

As a consequence, the major contributions of our work can be summarized into three-fold:
\begin{itemize}
 \item[$\bullet$] We manually construct a \textit{FineGPR-C} caption dataset for the first time to enable the semantic pretraining for Re-ID task.

 \item[$\bullet$] Based on it, a semantic-based pre-training approach named VTBR is proposed to learn visual representations from textual annotations on Re-ID event.
 \item[$\bullet$] Comprehensive experiments show that our VTBR method matches or exceeds the performance of existing methods for supervised or unsupervised pretraining on ImageNet with fewer images, which reveals the  applicability of semantic-based pretraining with new insights.
\end{itemize}

In the rest of the paper. we first review some related works of person re-identification methods and previous pre-training method in Section \ref{sec2}. Then in Section \ref{sec3}, we give more details about the first \textit{FineGPR-C} caption dataset, as well as learning procedure of the proposed VTBR pretraining method. Extensive evaluations compared with state-of-the-art methods and comprehensive analyses of the proposed approach are elaborated in Section \ref{sec4}. Conclusion and Future Works are given in Section \ref{sec5}.

\section{Related Works}
\label{sec2}
In this section, we have a brief review on the related work of person Re-ID method and pretraining approach.  The mainstream idea of the existing methods is to learn a robust model for feature representation.

\subsection{Person Re-ID Methods}
Actually, there are mainly two kinds of feature learning paradigms for person Re-ID tasks: (1) Hand-crafted based method and (2) Deep learning based approach, which are introduced as follow:

Traditional research works~\citep{farenzena2010person,zhao2014learning,yi2014deep} related to hand-crafted systems for person re-ID aim to design or learn discriminative representation or pedestrian features. For example, \cite{farenzena2010person} proposed
an appearance-based method for these situations where the number of candidates varies continuously. \cite{zhao2014learning} designed a novel approach of learning mid-level filters from automatically discovered patch clusters for person re-identification. Besides directly using mid-level color and texture features, some methods~\citep{yi2014deep} also strive to learn a similarity metric from image pixels directly, which can jointly learn the color feature, texture feature and metric in a unified manner. Unfortunately, these hand-crafted feature based models always fail to produce competitive results on large-scale datasets. The main reason is that these early works are mostly based on heuristic design, and thus they could not learn optimal discriminative features on current large-scale dataset.

Recently, benefited from the advances of deep neural networks and availability of large-scale datasets, person Re-ID performance in supervised learning has been significantly boosted to a new level~\citep{xiang2020multi,xiang2020unsupervised}, e.g. \cite{xiang2020multi} propose a feature fusion strategy based on traditional convolutional neural network with attention mechanism, which learns robust feature extraction and reliable metric learning in an end-to-end manner. \cite{lopez2020semantic} propose an end-to-end multi-modal CNN that combines image and context information with an attention module for scene recognition. \cite{ning2020feature} design a feature selection network that combines global and local fine-grained features for person Re-ID task. There is also study~\citep{ning2021jwsaa} that locate all the valuable areas of the features on the basis of the joint weak saliency mechanism and attention-aware model. Besides, some recent works~\citep{wei2018person,deng2018image} attempt to address unsupervised domain adaptation base on GAN (Generative Adversarial Network) model.
However, these approaches always require abundant computing resources to achieve satisfactory performance, and leveraging  GAN  network is unable to guarantee the quality of generated images.
Additionally, previous methods~\citep{pan2018two,chen2019mixed,park2020relation} either focus on designing various deep CNN structures to learn discriminative feature embeddings, or strive to explore better loss functions for deep neural network training~\citep{hermans2017defense,sun2020circle,varior2016gated,xiao2017margin}. In essence, these works always leverage models pretrained on ImageNet dataset to mitigate the shortage of person Re-ID data, which suffers from the limitation of large domain gap between ImageNet and person Re-ID data.

\subsection{Person Re-ID Pretraining}
ImageNet pretraining is a dominant paradigm in vision community. Recently, some researches have shown a surprising result that CNN-based pretraining on ImageNet has limited impacts on Re-ID system due to the large domain gap between ImageNet and person Re-ID data. On the one hand,
benefit from large-scale pedestrian datasets, researchers try to leverage large-scale Re-ID datasets to perform pretraining, which can help to learn a discriminative feature  representations with high quality. Specifically, \cite{fu2021unsupervised} proposed an unsupervised pretraining strategy with contrastive learning based on LUPerson dataset. \cite{yang2021unleashing} designed an unsupervised pre-training framework for Re-ID based on the contrastive learning.
\cite{luo2021self} proposed a self-supervised pre-training strategy for transformer-based person re-identification task, which can further reduce the domain gap and accelerate the pre-training.
On the other hand, to mine the potential of semantic features, several studies~\citep{radford2021learning,desai2021virtex} try to learn visual representations from textual annotations on basic computer vision tasks,
e.g., \cite{radford2021learning} demonstrated that learning visual models from natural language supervision is an efficient way to learn the state-of-the-art image representation, which has attracted considerable attention from both the academia and industry worldwide.
Since semantic feature is critical for representation learning, many research works~\citep{lin2019improving,jeong2021asmr} try to leverage the semantic feature to boost the performance of person Re-ID task, for example, \cite{lin2019improving} build an attribute-person recognition network to exploit both identity labels and attribute annotations for better semantic feature representation on person Re-ID task. \cite{jeong2021asmr} present an efficient and effective framework with semantic affinities for attribute-based person search. Unfortunately, these methods fail to achieve satisfactory performance due to the limited scale of textural annotations on person Re-ID event.
Inspired from these studies, in this paper, we manually construct a \textit{FineGPR-C} caption dataset for person Re-ID events, based on it, a pure semantic based Re-ID pretraining framework named \textbf{VTBR} is proposed to seek a common vision-language feature space with discriminative learning constraints for robust representation and better practical deployment.

Although VTBR inherits the structure of previous transformer-based CLIP~\citep{radford2021learning} and VirTex~\citep{desai2021virtex}, there exists some significant new designs in VTBR to allow it work for a very different manner: First, we only employ standard pretraining strategy in a supervised fashion with labeled image pairs, while CLIP adopts contrastive self-supervised learning with unlabeled image pairs from COCO dataset~\citep{lin2014microsoft}, and VirTex leverages the semantic density of captions to learn visual representations for downstream detection tasks, note that VirTex employs the dense caption of COCO Captions dataset (\textit{e.g.} five captions per image) to increase the image-caption pairs by five-fold, leading an expensive annotation costs in terms of worker hours, which indicate these works are more complex; Second, during the testing stage, both Text Encoder and Image Encoder are adopted for classification in previous CLIP, while our VTBR method only employs single visual backbone for downstream Re-ID task, allowing our method to be more flexible in real-world scenarios. Finally, this is the first time as far as we know, to significantly explore the potential of semantic features on Re-ID pretraining, from which we have proved that learning visual representation using textual annotations can be competitive to methods based on both supervised and unsupervised learning on ImageNet. We also hope that our dataset and method will shed light on some related researches to move forward, especially for semantic-based pretraining.

\section{Proposed Method}
\label{sec3}

\subsection{Problem Formulation}
For the pretraining of Re-ID task, given a labeled source dataset $S=\left\{x_{1}, x_{2}, \cdots, x_{N_{s}}\right\}$, consisting
of $N_{s}$ person images with manually annotated labels $Y=\left\{y_{1}, y_{2}, \cdots, y_{N_{s}}\right\}$, our goal is to learn a pretrained embedding function $\phi \left( \theta \right)$ from labeled source dataset $S$.  We also have an unlabeled target dataset $T=\left\{t_{1}, t_{2}, \cdots, t_{M}\right\}$. Note that there is non-overlapping in terms of identity between source domain and target domain in open set domain adaptation. Finally,
by leveraging both labeled source images from $S$ and unlabeled target samples in $T$, we can train a discriminative Re-ID model that generalized well in supervised or domain adaptive Re-ID task.

%------------------------
\begin{figure*}[!t]
\centering{\includegraphics[width=1.00\linewidth]{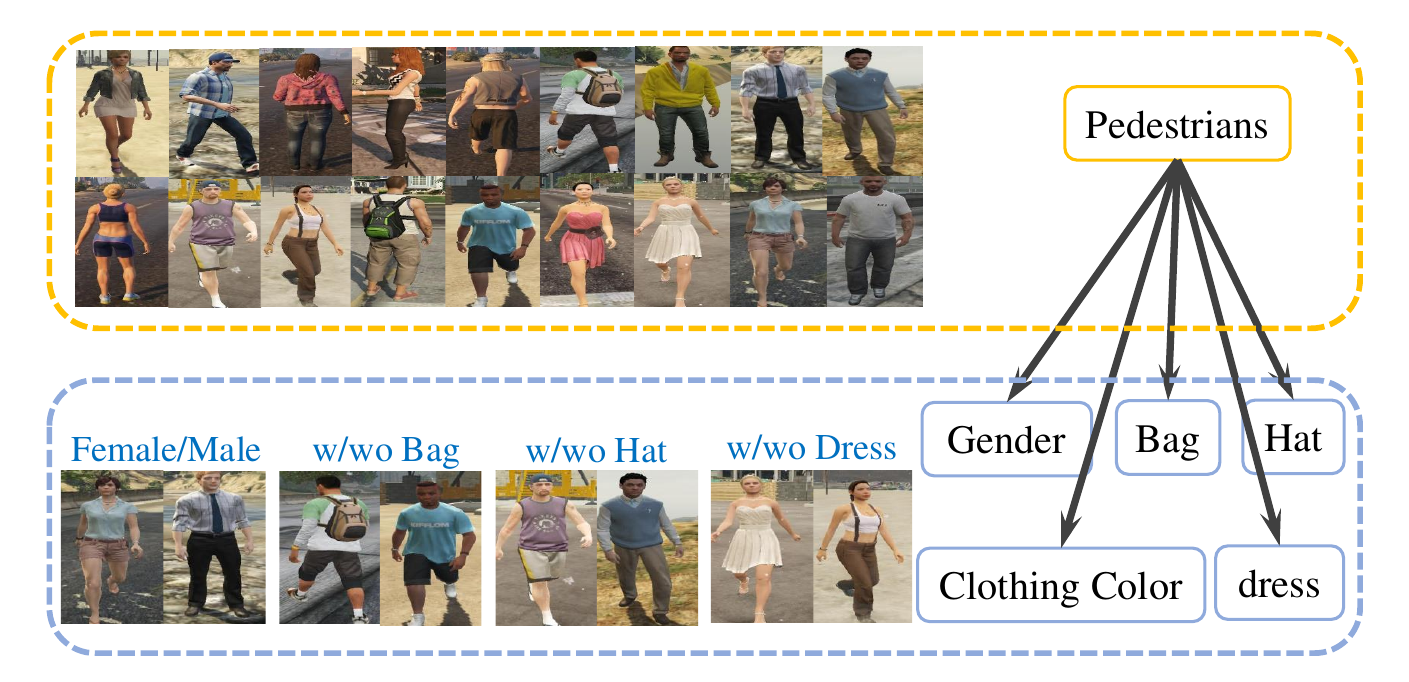}}
\caption{Attributes illustration at the identity level. We manually annotate these attribute at the identity level in a fine-grained manner.}
\label{fig2}
\end{figure*}
%-------------------------------------
%%------------------------
\begin{figure}[!t]
\centering{\includegraphics[width=0.85\linewidth]{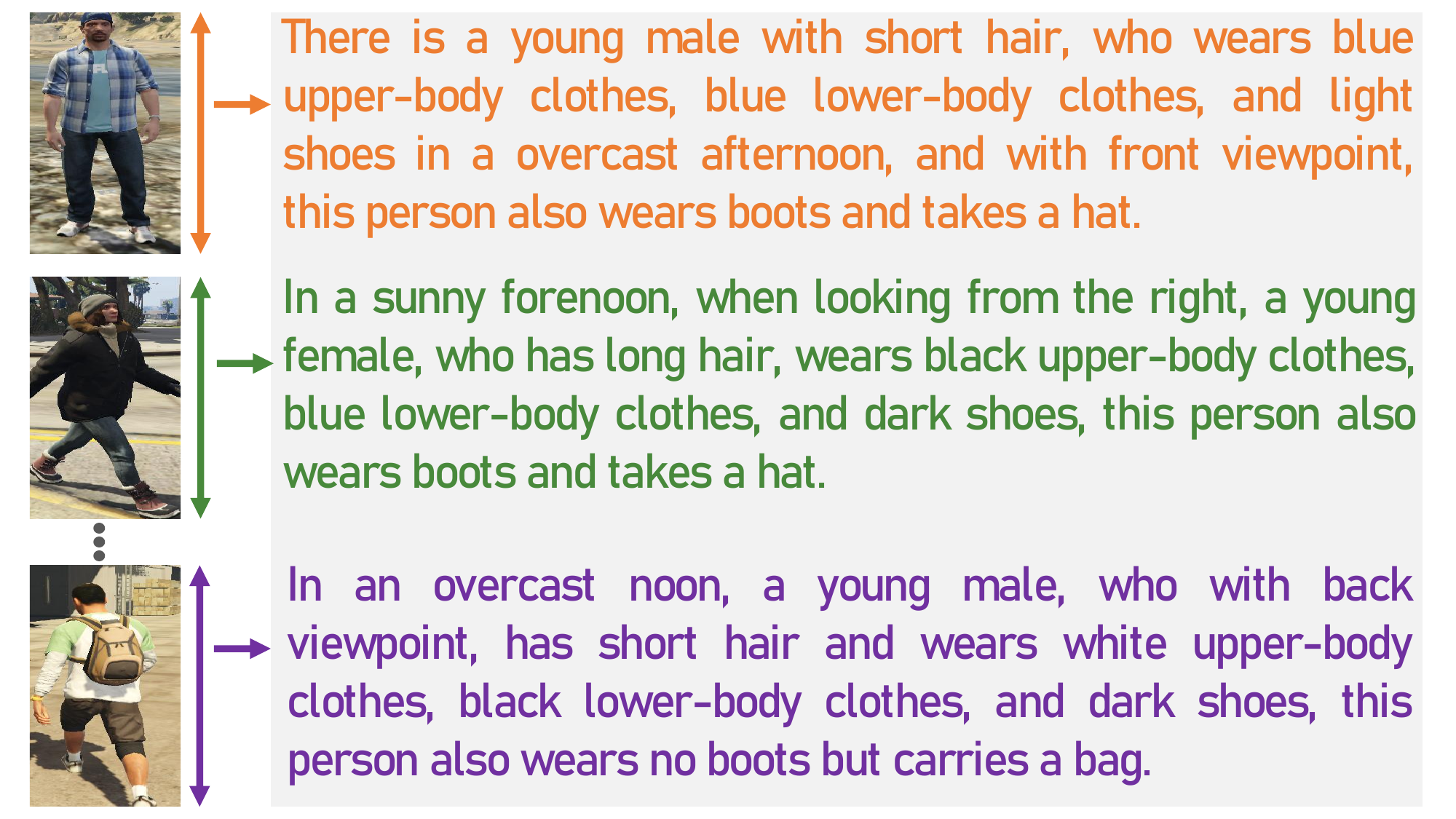}}
\caption{Some exemplars of semantic caption in our \textit{FineGPR-C} caption dataset, which is generated using our dynamic  caption generating strategy. }
\label{fig3}
\end{figure}
%%-------------------------------------

\subsection{\textit{FineGPR-C} Caption Dataset}
Data is the life-blood of training deep neural network models and ensuring their success. For the person Re-ID task, sufficient and high-quality data
are necessary for increasing the model's generalization capability. In this work, we ask the question: \textit{can we construct a person  dataset with captions which can be used as semantic-based pretraining on Re-ID task?} To answer this question, we revisit the previously developed \textit{FineGPR}~\citep{xiang2021less} dataset, which contains fine-grained attributes such as viewpoint, weather, illumination and background, as well as 13 accurate annotations at the identity level.  More details about attribute distribution of pedestrian can be available in Figure~\ref{fig2}.

%$\footnote[1]{\textcolor{black}{We only use one kind of \textit{FineGPR} caption in this work.}}$
%\subsection{\textit{FineGPR} Caption Generation}
To provide data foundation for semantic-based pretraining,
on the basis of \textit{FineGPR}, we introduce a dynamic strategy to generate high-quality captions with fine-grained attribute annotations for semantic-based pretraining. To be more specific, we rearrange the different attributes as word embeddings into caption expressions at the different position, and then generate semantically dense caption containing high-quality description, this gives rise to our newly-constructed \textit{FineGPR-C} caption dataset. Some exemplars of \textit{FineGPR-C} dataset are depicted in Figure~\ref{fig3}. It is worth mentioning that different pedestrian images have different captions by the different regular expressions.
%different images

\begin{figure*}[!t]
\centering{\includegraphics[width=1.00\linewidth]{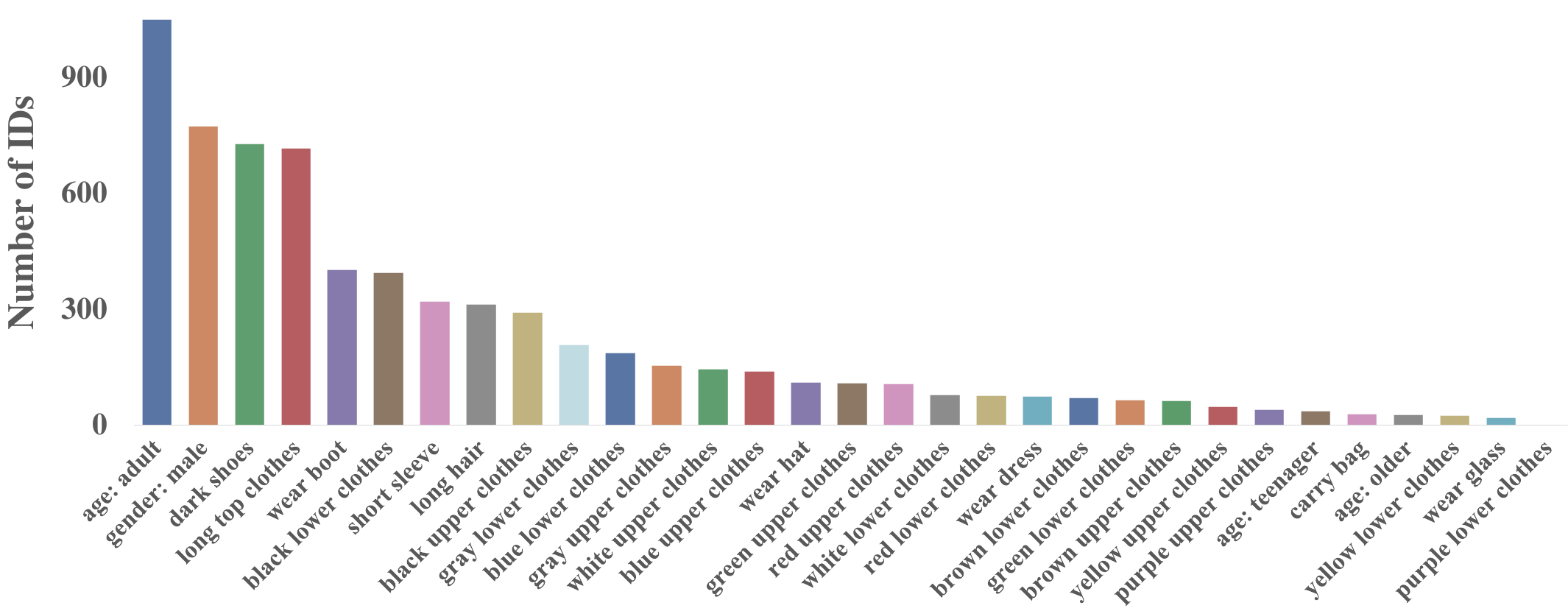}}
\caption{Distribution of attributes in terms of identity for \textit{FineGPR-C}  dataset. Please zoom in for the best view.}
\label{fig6}
\end{figure*}
%-------------------------------------

During the caption generation process, we found that there exists serious redundancy
among the different attributes in \textit{FineGPR-C}, especially for some attribute appears with larger probability. As shown in  Figure~\ref{fig6}, for example, the attribute of $adult$ in terms of Age accounts for more than three-quarters (nearly 1049/1150) of total identities in \textit{FineGPR-C} caption dataset, which may trigger a new problem that pedestrian attributes with high probability will degrade the diversity of generated caption dataset. To address this problem, we introduce a \textbf{R}efined \textbf{S}electing (\textbf{RS}) strategy to increase the inter-class diversity of different identities and minimize the intra-class variation of same identity. Particularly,
we set a threshold $\lambda$ to control the appearing probability of attributes in the final caption sentence $c$, which can dynamically select some representative pedestrian attributes who appears with a lower probability, finally the formula can be expressed as:
\begin{equation}
c = \left\{w_{1}, a_{1}, w_{2}, a_{2}, \cdots,  w_{K}, a_{K}\right\},\  if\ P_{a_{1}},  P_{a_{2}}, \cdots, P_{a_{K}} \leq \lambda
\label{eq1}
\end{equation}
where $K$ indicates the total number of identities, $w_{1}$, $w_{2}$, $\cdots$, $w_{K}$ denote fixed sentence words,  and $a_{1}$, $w_{2}$, $\cdots$, $a_{K}$ indicate labelled pedestrian attributes, respectively. $P_{a_{1}}$, $P_{a_{2}}$, $\cdots$, $P_{a_{K}}$ represents the appearing probability of the corresponding attribute annotation $a_{1}$, $w_{2}$, $\cdots$, $a_{K}$ in \textit{FineGPR}, respectively. In principle,
our overall goal for constructing \textit{FineGPR-C} caption dataset is to improve the caption's discriminative ability according to their attribute distribution, so the generated caption (token by token) will be more diversified and contain more discriminative information.
%More details about \textit{FineGPR} and \textit{FineGPR-C} can be found at our Project Homepage\footnote[1]{\url{https://github.com/JeremyXSC/FineGPR}}.

To this end, even though newly-built \textit{FineGPR-C} dataset is based on the previous \textit{FineGPR} dataset, \textit{FineGPR-C} is actually different from \textit{FineGPR} since it is constructed with fine-grained pedestrian attributes by our Refined Selecting (RS) strategy, which lays a data foundation for semantic pretraining framework VTBR. Please note that \textit{FineGPR-C} is also the first caption dataset for person Re-ID events, which will serve as a solid baseline in semantic pretraining and can greatly advance the research in Re-ID community.

%\begin{wrapfigure}[8]{l}[0em]{0.35\textwidth}%靠文字内容的左侧
%\includegraphics[width=0.15\textwidth]{images/IMG3.pdf}
%\caption{One visual example of semantically dense caption on GPR++ dataset.}
%\end{wrapfigure}

%------------------------
\begin{figure*}[!t]
\centering{\includegraphics[width=1.0\linewidth]{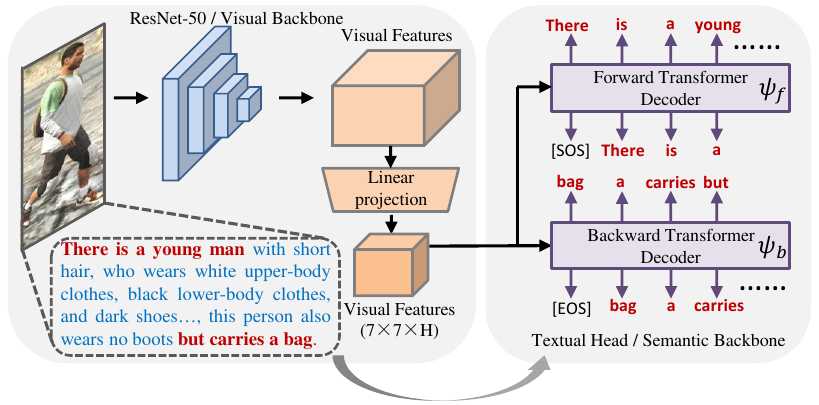}}
\caption{The framework of our vision-language supervised pretraining approach VTBR, which consists of a visual backbone ResNet-50 and
Transformer. The visual backbone extracts visual features, and transformers predict captions via bidirectional language modeling on the basis of visual features. After pretraining, the visual backbone is transferred to downstream Re-ID tasks.}
\label{fig4}
\end{figure*}
%-------------------------------------

\subsection{Our VTBR Approach}
In order to learn deep visual representations from textual annotations for Re-ID task, we introduce a semantic-based pretraining method VTBR based on our newly-built \textit{FineGPR-C} dataset. As illustrated in Figure~\ref{fig4},
our VTBR framework consists of a visual backbone ResNet-50~\citep{he2016deep} and semantic backbone Transformer~\citep{vaswani2017attention}, which extracts visual features of images and textual features of caption respectively. Firstly, the visual features extracted from ResNet-50 are used to predict captions of pedestrian images by transformer networks. Following the~\cite{desai2021virtex}, we use projection layer to receive features from the visual backbone, then put them to the textual head to predict captions with transformers for images, which adopts multiheaded self-attention both to propagate information of caption tokens and then provide a learning signal to the visual backbone during pretraining. Note that this projection layer is not used in downstream tasks. During training, we use the log-likelihood loss function to train the visual and semantic backbones in an end-to-end manner, which can be written as:
\begin{equation}
\mathcal{L}=\sum_{k=1}^{K+1} \log \left(p\left(T, V ; \psi_{f}, \phi\right)\right)+\sum_{k=0}^{K} \log \left(p\left(T, V ; \psi_{b}, \phi\right)\right)
\label{eq2}
\end{equation}
where $\psi_{f}$, $\psi_{b}$ and $\phi$ mean forward transformer, backward transformer and ResNet-50 respectively. $T$ and $V$ denote textual feature and visual feature separately. Log-probabilities are predicted by the linear layer of the last Transformer
layer over the token vocabulary. It is worth mentioning that both visual and semantic backbones are jointly trained to maximize the log-likelihood of the correct caption tokens.
Instead of adopting pretrained weight on ImageNet dataset,
we train our entire VTBR model from scratch on our \textit{FineGPR-C} caption dataset, whereas they rely on pretrained transformer to extract textual features.

%------------------------
\begin{figure*}[!t]
\centering{\includegraphics[width=1.0\linewidth]{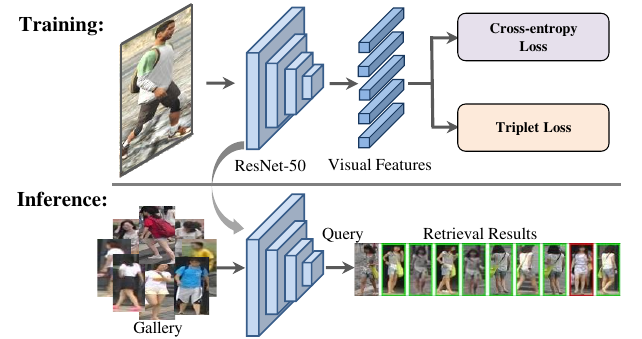}}
\caption{The illustration of training procedure for Re-ID tasks. First, the deep features of input images are extracted by CNN-based network, then, two commonly used loss functions (e.g., Cross-entropy loss and Triplet loss) are adopted for deep metric learning. Finally, the fine-tuned model is employed for Re-ID evaluation.}
\label{fig7}
\end{figure*}
%-------------------------------------

%Instead of predicting high-quality captions, our main goal is to
%learn transferable visual features which is beneficial to downstream Re-ID task.
%\begin{equation}
%\begin{aligned}
%\mathcal{L}(\theta, \phi) &=\sum_{t=1}^{T+1} \log \left(p\left(c_{t} \mid c_{0: t-1}, I ; \phi_{f}, \theta\right)\right) \\
%+& \sum_{t=0}^{T} \log \left(p\left(c_{t} \mid c_{t+1: T+1}, I ; \phi_{b}, \theta\right)\right)
%\end{aligned}
%\end{equation}

After obtaining the pretrained model based on our \textit{FineGPR-C} caption dataset, we perform downstream Re-ID evaluation\footnote[1]{In this work, we adopt a widely used open-source Re-ID backbone in~\cite{luo2019bag}, more details can be available at \url{https://github.com/michuanhaohao/reid-strong-baseline}.} continuously. Specifically, we adopt global features extracted by visual backbone ResNet-50 to perform metric learning, the pipeline for Re-ID task is illustrated in Figure~\ref{fig7}. It is worth mentioning that we only modify the output dimension of the latest fully-connected layer to the number of training identities~\citep{xiang2021taking}. During the period of testing, we extract the 2,048-dim pool-5 vector for retrieval under the Euclidean distance.

\section{Experimental Results}
\label{sec4}
\subsection{Datasets}
\label{sec4.1}
In this paper, we conduct experiments on three large-scale
public datasets, which include Market-1501~\citep{zheng2015scalable}, DukeMTMC-reID~\citep{ristani2016performance,zheng2017unlabeled} and CUHK03~\citep{li2014deepreid} datasets.

\textbf{Market-1501}~\citep{zheng2015scalable} contains 32,668 labeled images of 1,501 identities captured from campus in Tsinghua University.  Each identity is captured by at most 6 cameras. The training set contains 12,936 images from 751 identities and the test set contains 19,732 images from 750 identities.

\textbf{DukeMTMC-reID}~\citep{ristani2016performance,zheng2017unlabeled} is collected from Duke University with 8 cameras, it has 36,411 labeled images belonging to 1,404 identities and contains 16,522 training images from 702 identities, 2,228 query images from another 702 identities and 17,661 gallery images.

%\footnote[1]{Note that the DukeMTMC and its derived datasets have been officially removed due to some ethic concerns. Here we include it only for the sake of comparison to some existing results. We discourage further usage of DukeMTMC datasets in the future.}

\textbf{CUHK03}~\citep{li2014deepreid} contains 14,097 images of 1,467 identities.
Following the CUHK03-NP protocol~\citep{zhong2017re}, it is divided into 7,365 images of 767 identities as the training set, and the remaining 6,732 images of 700 identities as the testing set.

In our experiments, we follow the standard evaluation protocol~\citep{zheng2015scalable} used in Re-ID task , and adopt mean Average Precision (mAP) and Cumulative Matching Characteristics (CMC) at rank-1 and rank-5 for performance evaluation on downstream re-ID task.

\subsection{Implementation Details}
For the pretraining of VTBR, we apply standard random cropping and normalization as data
augmentation. Following the training procedure in~\citep{desai2021virtex}, we adopt SGD with momentum 0.9 and weight decay $10^{-4}$ wrapped in LookAhead~\citep{kukunuri2019lookahead} with $\alpha$=0.5 and 5 steps. We empirically set the $\lambda$=0.8 in Eq.~\ref{eq1}.
The max learning rate of visual backbone is $2\times10^{-1}$; learning rate of the textual head is set as $1\times10^{-3}$.
For the downstream Re-ID task, we closely follow a widely used open-source project~\citep{luo2019bag} as standard baseline, which is built only with commonly used softmax cross-entropy loss~\citep{zhang2018generalized} and triplet loss~\citep{hermans2017defense}
on vanilla ResNet-50 with no bells and whistles. Following the practice in~\citep{luo2019bag}, the batch size of training samples is set as 64. As for triplet selection, we randomly selected 16 persons and sampled 4 images for each identity, m is set as 0.5  as triplet margin.  Adam method and warmup learning strategy are also adopted to optimize the model. All the experiments are performed on PyTorch~\citep{paszke2019pytorch} with two Nvidia GeForce RTX 3090 GPUs on a server equipped with a Intel Xeon Gold 6240 CPU.

%------------------------
\begin{figure*}[!t]
\centering{\includegraphics[width=0.75\linewidth]{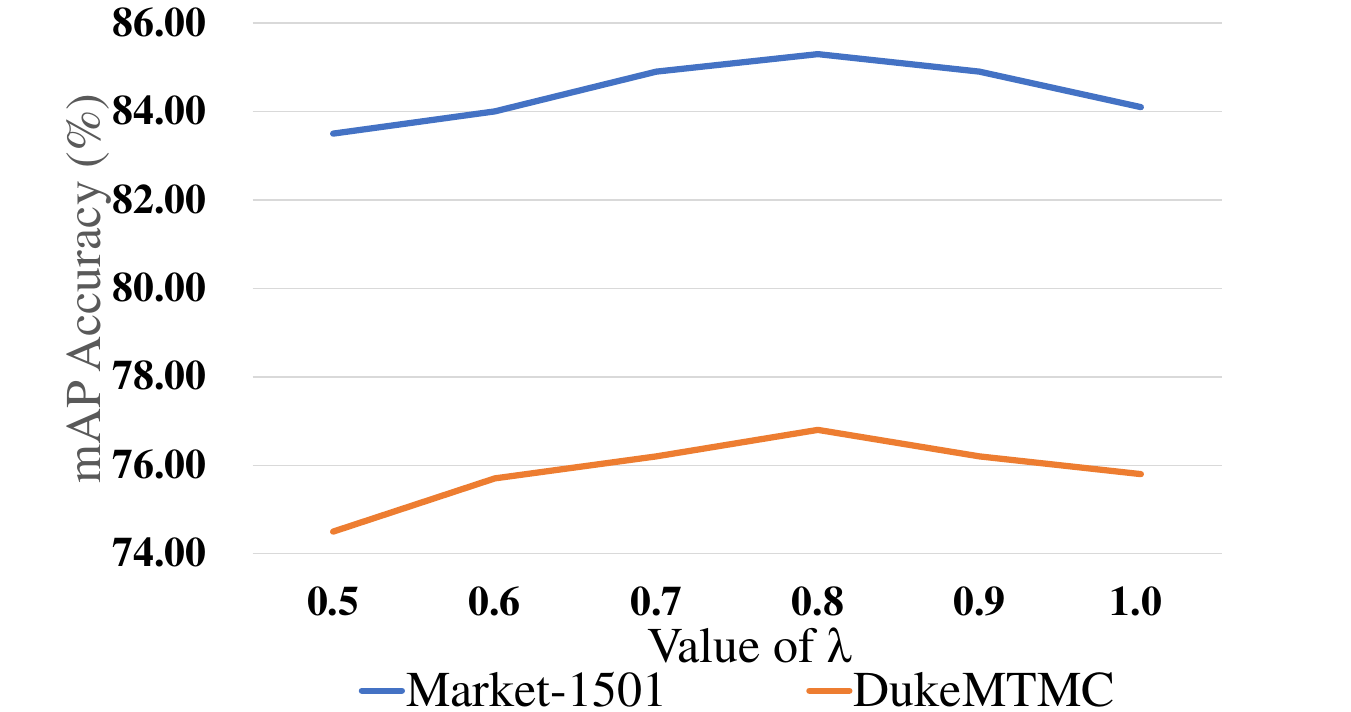}}
\caption{Sensitivity of our semantic-based pretraining  VTBR to key parameter $\lambda$ on supervised Re-ID tasks.}
\label{fig8}
\end{figure*}
%-------------------------------------

\subsection{Important Parameter}
In this section, we evaluate the impacts of parameter $\lambda$ in Eq.~\ref{eq1}, which controls the probability of attributes appears in the final caption sentence $c$. As depicted in Figure~\ref{fig8}, it can be easily observed that when $\lambda$ is small, the performance is not optimal because the selected attribute is way too limited to a very small portion, and thus our VTBR could not mine the relevant image regions or discriminative parts to learn semantic features. The $\lambda$ should also not be set too large, otherwise the performance drop dramatically since generated caption can not maintain a high diversity. Specially, $\lambda=0.8$ yields the best accuracy for Re-ID task.

%In downstream re-ID task, we fine-tune the visual backbone end-to-end on Re-ID dataset.
\begin{table}[!t]
  \centering
  \caption{Comparisons between traditional CNN-based and semantic-based VTBR  pretraining on supervised Re-ID tasks. ``S" denotes semantic feature. For different pretraining method, ``ResNet (ImageNet)" means we pretrain ResNet-50 on ImageNet dataset, same for ResNet-50 (\textit{FineGPR}),  VTBR (\textit{FineGPR-C}) and VTBR+RS (\textit{FineGPR-C}). \textcolor[rgb]{1.00,0.39,0.09}{\textbf{Orange}} indicates the best and \textcolor[rgb]{0.20,0.40,0.80}{\textbf{Blue}} the second best.}
  \small
  \setlength{\tabcolsep}{1.0mm}{
    \begin{tabular}{lcccccccc}
    \toprule
    \multicolumn{3}{c}{Supervised Fine-tuning $\rightarrow$}  & \multicolumn{2}{c}{Market-1501} & \multicolumn{2}{c}{DukeMTMC}  & \multicolumn{2}{c}{CUHK03} \\
    \cmidrule(lr){1-3}  \cmidrule(lr){4-5} \cmidrule(lr){6-7}  \cmidrule(lr){8-9}
    Pretrain $\downarrow$  &S  &\#Imgs   & Rank-1 & mAP   & Rank-1 & mAP  & Rank-1 & mAP\\
    \toprule
    ResNet (ImageNet)  & $\times$  & 1.28M  & \textcolor[rgb]{0.20,0.40,0.80}{\textbf{94.3}}  & \textcolor[rgb]{0.20,0.40,0.80}{\textbf{85.0}}  & \textcolor[rgb]{0.20,0.40,0.80}{\textbf{86.7}}  & \textcolor[rgb]{0.20,0.40,0.80}{\textbf{76.6}}  & 61.6  & \textcolor[rgb]{0.20,0.40,0.80}{\textbf{59.0}}  \\
    ResNet (\textit{FineGPR})  & $\times$  & 2.00M  & 85.5  & 74.2  & 63.9  & 59.8  & 43.3  & 37.8  \\
    VTBR (\textit{FineGPR-C})  & \checkmark  & 1.83M  & 93.6  & 83.7  & 85.0  & 72.9  & \textcolor[rgb]{0.20,0.40,0.80}{\textbf{61.8}}  & 58.6  \\
    VTBR+RS(\textit{FineGPR-C})  & \checkmark  & 0.91M  & \textcolor[rgb]{1.00,0.39,0.09}{\textbf{94.9}}  & \textcolor[rgb]{1.00,0.39,0.09}{\textbf{85.3}}  & \textcolor[rgb]{1.00,0.39,0.09}{\textbf{87.3}}  & \textcolor[rgb]{1.00,0.39,0.09}{\textbf{76.8}}  & \textcolor[rgb]{1.00,0.39,0.09}{\textbf{61.9}}  & \textcolor[rgb]{1.00,0.39,0.09}{\textbf{59.3}}  \\
    \bottomrule
    \end{tabular}}%
  \label{tab1}%
\end{table}%

\subsection{Supervised Fine-tuning}
In this work, the caption data for Re-ID event is the fundamental part of the semantic-based pretraining baseline. Here, we adopt supervised fine-tuning performance on real datasets
as the indicator to show the quality of \textit{FineGPR-C} caption dataset.  From Table~\ref{tab1}, we
can obviously observe that the results of supervised learning are significantly promoted by using our method. For example, when training and testing on Market-1501 with ImageNet pretrained model, we can only achieve a rank-1 accuracy of \textbf{94.3\%}, while our VTBR method on \textit{FineGPR-C} can obtain a competitive performance of \textbf{93.6\%}. After employing the Refined Selecting strategy, our VTBR+RS reaches a remarkable performance of \textbf{94.9\%} with \textbf{1.4$\times$} fewer pretraining images (\textcolor[rgb]{1.00,0.39,0.09}{\textbf{0.91M}} vs. \textcolor[rgb]{0.20,0.40,0.80}{\textbf{1.28M}}), leading to a record mAP performance of \textbf{85.3\%}.  Not surprisingly, same performance gain can also be achieved on DukeMTMC-reID dataset. The success of VTBR can be largely contributed to the discriminative features learned by semantic captions in a data-efficient manner.

\begin{table*}[!t]
  \centering
  \caption{Comparisons between traditional CNN-based and semantic-based VTBR pretraining on domain adaptive Re-ID tasks. ``S" denotes semantic feature. For different pretraining method, ``ResNet (ImageNet)" means we pretrain ResNet-50 on ImageNet dataset, same for ResNet-50 (\textit{FineGPR}),  VTBR (\textit{FineGPR-C}) and VTBR+RS (\textit{FineGPR-C}). \textcolor[rgb]{1.00,0.39,0.09}{\textbf{Orange}} indicates the best and \textcolor[rgb]{0.20,0.40,0.80}{\textbf{Blue}} the second best.}
  \small
  \setlength{\tabcolsep}{0.71mm}{
    \begin{tabular}{lcccccccc}
    \toprule
     \multicolumn{3}{c}{Domain Adaptation $\rightarrow$}  & \multicolumn{3}{c}{DukeMTMC$\rightarrow$Market} & \multicolumn{3}{c}{Market$\rightarrow$DukeMTMC} \\
    \cmidrule(lr){1-3}  \cmidrule(lr){4-6} \cmidrule(lr){7-9}
    Pretrain $\downarrow$  &S  &\#Imgs  & Rank-1 & Rank-5  & mAP   & Rank-1 & Rank-5  & mAP  \\
    \toprule
    ResNet (ImageNet)  & $\times$  & 1.28M  & \textcolor[rgb]{0.20,0.40,0.80}{\textbf{48.0}}  & 64.1  & \textcolor[rgb]{0.20,0.40,0.80}{\textbf{21.7}}  & \textcolor[rgb]{1.00,0.39,0.09}{\textbf{24.5}}  & \textcolor[rgb]{1.00,0.39,0.09}{\textbf{38.8}}  & \textcolor[rgb]{1.00,0.39,0.09}{\textbf{13.8}}  \\
    ResNet (\textit{FineGPR})  & $\times$  & 2.00M  & 44.2  & 62.8  & 20.5  & 20.8  & 33.5  & 10.2 \\
   VTBR (\textit{FineGPR-C})  & \checkmark  & 1.83M  & 45.9  & \textcolor[rgb]{0.20,0.40,0.80}{\textbf{64.8}}  & 21.2  & 21.3  & 34.6  & 10.9 \\
    VTBR+RS (\textit{FineGPR-C})   & \checkmark  & 0.91M  & \textcolor[rgb]{1.00,0.39,0.09}{\textbf{50.6}}  & \textcolor[rgb]{1.00,0.39,0.09}{\textbf{67.7}}  & \textcolor[rgb]{1.00,0.39,0.09}{\textbf{23.8}}  &  \textcolor[rgb]{0.20,0.40,0.80}{\textbf{24.3}}  & \textcolor[rgb]{0.20,0.40,0.80}{\textbf{38.4}}  & \textcolor[rgb]{0.20,0.40,0.80}{\textbf{13.5}}   \\
    \bottomrule
    \end{tabular}}%
  \label{tab2}%
\end{table*}%

\subsection{Unsupervised Domain Adaption}
Our semantic-based pretraining method enjoys the benefits of flexible corner scenarios of domain adaptive Re-ID tasks, where labelled data in target domain is hard to obtain. In this section, we present four domain adaptive Re-ID tasks on several benchmark datasets.  More detailed results can be seen in Table~\ref{tab2} and Table~\ref{tab2_2}. For instance, when trained on DukeMTMC-reID dataset,  it can be easily observed that our VTBR+RS achieves a significant rank-1 performance of \textbf{50.6\%} and \textbf{5.7\%} on Market-1501 and CUHK03 respectively, outperforming the ImageNet pretraining by \textbf{+2.6\%} and \textbf{+0.8\%} in terms of rank-1 accuracy. When trained on Market-1501 dataset, our method can also lead to an obvious improvement of \textbf{+1.9\%} on CUHK03 in rank-1 accuracy. However, when tested on DukeMTMC-reID dataset, it is surprising to find that our method obtain a slightly inferior performance than ImageNet pretraining
 (mAP \textcolor[rgb]{0.20,0.40,0.80}{\textbf{13.5\%}} vs. \textcolor[rgb]{1.00,0.39,0.09}{\textbf{13.8\%}}, \textcolor[rgb]{0.20,0.40,0.80}{\textbf{0.91M}} vs. \textcolor[rgb]{1.00,0.39,0.09}{\textbf{1.28M}} images). We suspect that captions generated on \textit{FineGPR} have obvious domain gap with DukeMTMC-reID dataset since there are some occlusion and multiple persons in the queries, which will undoubtedly degrade the performance of our method.
% \textcolor[rgb]{0.20,0.40,0.80}{\textbf{0.91M}} vs. \textcolor[rgb]{1.00,0.39,0.09}{\textbf{1.28M}}
%\textcolor[rgb]{1.00,0.39,0.09}{\textbf{0.91M}}

\begin{table*}[!t]
  \centering
  \caption{Comparisons between traditional CNN-based and semantic-based VTBR pretraining on domain adaptive Re-ID tasks. ``S" denotes semantic feature. For different pretraining method, ``ResNet (ImageNet)" means we pretrain ResNet-50 on ImageNet dataset, same for ResNet-50 (\textit{FineGPR}),  VTBR (\textit{FineGPR-C}) and VTBR+RS (\textit{FineGPR-C}). \textcolor[rgb]{1.00,0.39,0.09}{\textbf{Orange}} indicates the best and \textcolor[rgb]{0.20,0.40,0.80}{\textbf{Blue}} the second best.}
  \small
  \setlength{\tabcolsep}{0.57mm}{
    \begin{tabular}{lcccccccc}
    \toprule
     \multicolumn{3}{c}{Domain Adaptation $\rightarrow$}  & \multicolumn{3}{c}{DukeMTMC$\rightarrow$CUHK03} & \multicolumn{3}{c}{Market$\rightarrow$CUHK03} \\
    \cmidrule(lr){1-3}  \cmidrule(lr){4-6} \cmidrule(lr){7-9}
    Pretrain $\downarrow$  &S  &\#Imgs  & Rank-1 & Rank-5  & mAP   & Rank-1 & Rank-5  & mAP \\
    \toprule
    ResNet (ImageNet)  & $\times$  & 1.28M  & \textcolor[rgb]{0.20,0.40,0.80}{\textbf{4.9}}  & \textcolor[rgb]{0.20,0.40,0.80}{\textbf{11.6}}  & \textcolor[rgb]{0.20,0.40,0.80}{\textbf{5.6}}  & 3.9  & 8.6  & 4.0  \\
    ResNet (\textit{FineGPR})  & $\times$  & 2.00M  & 4.7  & 11.3  & 5.5  & 4.5  & 11.5  & 4.3 \\
   VTBR (\textit{FineGPR-C})  & \checkmark  & 1.83M  & 4.9  & 11.4  & 5.2  & \textcolor[rgb]{0.20,0.40,0.80}{\textbf{5.4}}  & \textcolor[rgb]{0.20,0.40,0.80}{\textbf{11.8}}  & \textcolor[rgb]{0.20,0.40,0.80}{\textbf{5.8}} \\
    VTBR+RS (\textit{FineGPR-C})   & \checkmark  & 0.91M  & \textcolor[rgb]{1.00,0.39,0.09}{\textbf{5.7}}  & \textcolor[rgb]{1.00,0.39,0.09}{\textbf{13.1}}  & \textcolor[rgb]{1.00,0.39,0.09}{\textbf{5.7}}  & \textcolor[rgb]{1.00,0.39,0.09}{\textbf{5.8}}  & \textcolor[rgb]{1.00,0.39,0.09}{\textbf{13.2}}  & \textcolor[rgb]{1.00,0.39,0.09}{\textbf{6.2}} \\
    \bottomrule
    \end{tabular}}%
  \label{tab2_2}%
\end{table*}%

\begin{table}[!t]
  \centering
  \caption{Performance comparison with other baselines of CNN architecture on supervised Re-ID tasks. * represents the attention-based method. \textcolor[rgb]{1.00,0.39,0.09}{\textbf{Orange}} indicates the best and \textcolor[rgb]{0.20,0.40,0.80}{\textbf{Blue}} the second best.}
  \small
  \setlength{\tabcolsep}{0.97mm}{
    \begin{tabular}{lcccccc}
    \toprule
    \multirow{2}[4]{*}{Methods} & \multicolumn{2}{c}{Market-1501} & \multicolumn{2}{c}{DukeMTMC} & \multicolumn{2}{c}{CUHK03} \\
\cmidrule(lr){2-3}  \cmidrule(lr){4-5}  \cmidrule(lr){6-7}      & Rank-1 & mAP   & Rank-1 & mAP   & Rank-1 & mAP \\
    \midrule
    BoW+XQDA~\citep{zheng2015scalable} & 44.4  & 20.8  & 25.1  & 12.2  & 6.4   & 6.4  \\
    LOMO+XQDA~\citep{liao2015person} & 43.8  & 22.2  & 30.8  & 17.0  & 12.8  & 11.5  \\
    SVDNet~\citep{sun2017svdnet} & 82.3  & 62.1  & 76.7  & 56.8  & 41.5  & 37.2  \\
    CASN(IDE)$^{*}$~\citep{zheng2019re} & 92.0  & 78.0  & \textcolor[rgb]{0.20,0.40,0.80}{\textbf{84.5}}  & 67.0  & 57.4  & 50.7  \\
    DaRe~\citep{wang2018resource}  & 86.4  & 69.3  & 75.2  & 57.3  & 55.1  & 51.3  \\
    PCB+RPP$^{*}$~\citep{sun2018beyond} & \textcolor[rgb]{0.20,0.40,0.80}{\textbf{93.8}}  & \textcolor[rgb]{0.20,0.40,0.80}{\textbf{81.6}}  & 83.3  & \textcolor[rgb]{0.20,0.40,0.80}{\textbf{69.2}}  & \textcolor[rgb]{1.00,0.39,0.09}{\textbf{63.7}}  & \textcolor[rgb]{0.20,0.40,0.80}{\textbf{57.5}}  \\
    \midrule
    VTBR+RS (Ours) & \textcolor[rgb]{1.00,0.39,0.09}{\textbf{94.9}}  & \textcolor[rgb]{1.00,0.39,0.09}{\textbf{85.3}}  & \textcolor[rgb]{1.00,0.39,0.09}{\textbf{87.3}}  & \textcolor[rgb]{1.00,0.39,0.09}{\textbf{76.8}}  & \textcolor[rgb]{0.20,0.40,0.80}{\textbf{61.9}}  & \textcolor[rgb]{1.00,0.39,0.09}{\textbf{59.3}}  \\
    \bottomrule
    \end{tabular}}%
  \label{tab3}%
\end{table}%

\subsection{Comparison with Other Methods}
In this section, we compare our results with existing methods of CNN architecture in Table~\ref{tab3}. Note that we do not apply any post-processing method like Re-Rank~\citep{zhong2017re} in our approach. As we can see, we can achieve state-of-the-art performance on Market-1501 and DukeMTMC-reID dataset with considerable advantages, by simply applying our semantic-based pre-training strategy, we can obtain a remarkable mAP performance of \textbf{85.3\%} and \textbf{76.8\%} on Market-1501 and DukeMTMC dataset respectively, leading a significant improvement of \textbf{+3.7\%} and \textbf{+7.6\%} when compared with second best method PCB+RPP~\citep{sun2018beyond}. Surprisingly, we also find an interesting phenomenon that performance of  VTBR+RS is slightly inferior and less competitive compared with PCB+RPP~\citep{sun2018beyond} on CUHK03 dataset (Rank-1 \textcolor[rgb]{1.00,0.39,0.09}{\textbf{63.7\%}} vs. \textcolor[rgb]{0.20,0.40,0.80}{\textbf{61.9\%}}). It is probably because that the attention or within-part consistency among image of same pedestrian in CUHK03 dataset can produce principled supervisory signals to baseline CNN architecture, which can greatly enhance the pedestrian discriminability of attention-based model on Re-ID task.

\subsection{Visualization}
In order to verify the effectiveness of our proposed VTBR method, we show more qualitative examples of Grad-CAM~\citep{muhammad2020eigen} visualizations in Figure~\ref{fig5}. Compared with ImageNet pretraining method, we observe that our model attends to relevant image regions or discriminative parts for making caption predictions, indicating that VTBR can greatly help the model learn more global context information and meaningful visual features with better semantic understanding, which significantly makes our semantic-based pretraining VTBR model more robust to perturbations.

%learn more global context information and more discriminative parts, which makes the model more robust to perturbations.

%------------------------
\begin{figure}[!t]
\centering{\includegraphics[width=\linewidth]{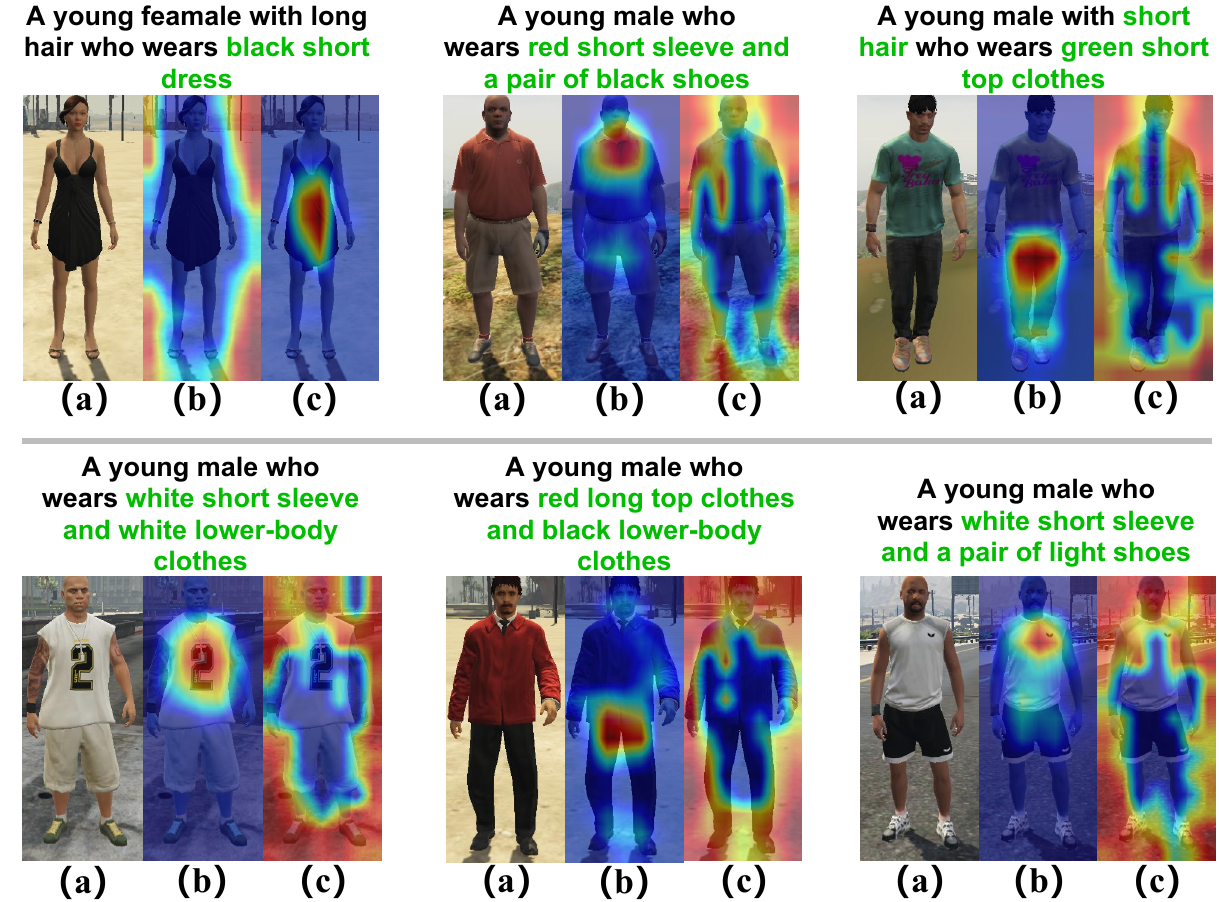}}
\caption{Visualization of attention maps with Grad-CAM~\citep{muhammad2020eigen}. (a) Original images, (b) CNN-based pretraining method on ImageNet, (c) Our semantic-based VTBR method on \textit{FineGPR-C} caption dataset. It can be easily observed that semantic pretraining method can capture global context information and more discriminative parts, which are further enhanced in our proposed VTBR method for better performance.}
\label{fig5}
\end{figure}
%-------------------------------------

\subsection{Discussion}
According to the experiments from Table~\ref{tab1}, Table~\ref{tab2} and Table~\ref{tab2_2}, semantic-based pretraining strategy demonstrates its competitiveness and priority since it can bridge the gap beween ImageNet and downstream person Re-ID data. Despite its promising performance on person Re-ID, we note that there are several limitations in our VTBR method.

First, the premise of our semantic-based VTBR method is that caption dataset for downstream task is required to be constructed based on domain-specific knowledge and  have a high diversity. However, there exists a fact that different  vision tasks have different caption paradigm, this means the caption generation process may be repeated by carefully designed strategy when downstream task makes a change in different scenarios;

Second, the generated caption dataset \textit{FineGPR-C} may still have domain gap with some collected datasets (e.g. DukeMTMC-reID) in terms of viewpoint, weather, illumination,  as well as background, which may bring negative impacts and degrade the performance for downstream Re-ID task to a certain extent.
We believe addressing these challenges are promising direction of our work for future research.

%% Table generated by Excel2LaTeX from sheet 'Sheet1'
%\begin{table}[t]
%  \centering
%  \caption{Performance (\%) comparisons between ImageNet pretraining and  \textit{GPR+} pretraining in domain adaptation manner.}
%  \small
%  \setlength{\tabcolsep}{0.8mm}{
%    \begin{tabular}{ccccccc}
%    \toprule
%    Testing $\rightarrow$  & \multicolumn{3}{c}{GPR+$\rightarrow$Market} & \multicolumn{3}{c}{GPR+$\rightarrow$DukeMTMC} \\
%    \midrule
%    Pretrain $\downarrow$ & Rank-1 & Rank-5 & mAP   & Rank-1 & Rank-5 & mAP \\
%    \toprule
%    ImageNet & 44.2  & 62.2  & 21.9  & 25.0  & 39.4  & 15.4  \\
%    GPR+  & 37.1  & 54.5  & 19.8  & 30.0  & 43.2  & 14.7  \\
%    GPR+(refined) & 37.8  & 55.4  & 20.1  & 30.1  & 42.5  & 13.3  \\
%    \bottomrule
%    \end{tabular}}%
%  \label{tab3}%
%\end{table}%

\section{Conclusion and Future Work}
\label{sec5}
This paper takes a big step forward to rethink person re-identification via semantic-based pretraining. Specially, we construct the first \textit{FineGPR-C} caption dataset for person Re-ID events, which covers human describing in a fine-grained manner.
Based on it, we present a simple yet effective semantic-based pretraining method to replace the ImageNet pretraining, which
helps to learn visual representations from textual annotations on downstream Re-ID task.
Extensive experiments conducted on several benchmarks show that our method outperforms the traditional ImageNet pretraining -- both in supervised and unsupervised manner -- by a clear margin, revealing the potential of semantic-based pretraining for further studies. In the future, we will focus on other downstream vision tasks with semantic-based VTBR, such as human parts segmentation and pose estimation.

\bmhead{Acknowledgments}

This work was supported by the National Natural Science Foundation of China under Grant No. 61977045 and No. 81974276.
The authors would like to thank the anonymous reviewers for their valuable suggestions and constructive criticisms.

\section*{Declarations}

\begin{itemize}
\item \textbf{Funding} \\  This work was partially supported by the National Natural Science Foundation of China under Grant No. 61977045 and No. 81974276.
\item \textbf{Conflict of interest} \\  The authors declare that they have no conflict of interest.
\item \textbf{Ethics approval} \\  All procedures performed in studies involving human participants were in accordance with
the ethical standards of the institutional and/or national research committee.
\item \textbf{Consent to participate} \\  All human participants consented for participating in this study.
\item \textbf{Consent for publication} \\  All contents in this paper are consented for publication.
\item \textbf{Availability of data and material} \\  The data used for the experiments in this paper are available online, see Section~\ref{sec4.1} for more details.
\item \textbf{Code availability} \\  The Re-ID baseline implementation is open-source; The dataset is also publicly available at \url{https://github.com/JeremyXSC/FineGPR}.
\item \textbf{Authors' contributions} \\  Suncheng Xiang, and Yuzhuo Fu contributed conception and design of the study. Jingsheng Gao, Zirui Zhang, Mengyuan Guan and Binjie Yan contributed to experimental process and evaluated and interpreted model results. Yuzhuo Fu and Dahong Qian obtained funding for the project. Ting Liu, Dahong Qian and Yuzhuo Fu provided clinical guidance. Suncheng Xiang drafted the manuscript. All authors contributed to manuscript revision, read and approved the submitted version.
\end{itemize}

%\bmhead{Funding}  This work was supported by the National Natural Science Foundation of China under Project (Grant No. 61977045).
%
%\bmhead{Conflict of interest}  The authors declare that they have no conflict of interest.
%
%\bmhead{Ethics approval}  All procedures performed in studies involving human participants were in accordance with
%the ethical standards of the institutional and/or national research committee.
%
%\bmhead{Consent to participate}  All human participants consented for participating in this study.
%
%\bmhead{Consent for publication}  All contents in this paper are consented for publication.
%
%\bmhead{Availability of data and material}  The data used for the experiments in this paper are available online, see Section~\ref{sec4.1} for more details.
%
%\bmhead{Code availability}  The Re-ID baseline implementation is open-source; The dataset is also publicly available at \url{https://github.com/JeremyXSC/FineGPR}.
%
%\bmhead{Authors' contributions}  Suncheng Xiang, and Yuzhuo Fu contributed conception and design of the study. Jingsheng Gao, Zirui Zhang, Mengyuan Guan and Binjie Yan contributed to experimental process and evaluated and interpreted model results. Yuzhuo Fu obtained funding for the project. Ting Liu and Yuzhuo Fu provided clinical guidance. Suncheng Xiang drafted the manuscript. All authors contributed to manuscript revision, read and approved the submitted version.

\bibliography{sn-bibliography}% common bib file


\end{document}